\documentclass[11pt,a4paper]{article}
\usepackage[hyperref]{emnlp-ijcnlp-2019}
\usepackage[export]{adjustbox}
\usepackage{graphicx}
\usepackage{subfig}
\usepackage{times}
\usepackage{latexsym}
\usepackage{url}
\usepackage{pict2e}
\usepackage{drs}
\usepackage{algorithm}
\usepackage{amsmath, amssymb}
\usepackage{multirow}
\usepackage{multicol}
\usepackage[noend]{algpseudocode}
\usepackage{amsthm}
\usepackage{textcomp}
\usepackage{starfont}

\newcommand{\ignore}[1]{}
\theoremstyle{plain}

\usepackage{tikz}
\usetikzlibrary{arrows,positioning}

\aclfinalcopy % Uncomment this line for the final submission
\title{Semantic Graph Parsing with Recurrent Neural Network DAG Grammars}

\author{Federico Fancellu$^{\textrm\Sun}$ \and Sorcha Gilroy$^{\textrm\Moon}$ \and Adam Lopez$^{\textrm\Terra}$ \and Mirella Lapata$^{\textrm\Terra}$\\
$^{\textrm\Sun}$Samsung AI Centre (SAIC), Toronto, Canada\\
$^{\textrm\Moon}$Peak AI, Manchester, United Kingdom\\
$^{\textrm\Terra}$University of Edinburgh, Edinburgh, United Kingdom}

\date{}

\makeatletter
\newcommand{\thickhline}{%
    \noalign {\ifnum 0=`}\fi \hrule height 1pt
    \futurelet \reserved@a \@xhline
}
\makeatother

\begin{document}
\maketitle
\begin{abstract}
 Semantic parses are directed acyclic graphs (DAGs), so semantic parsing should be modeled as graph prediction. But predicting graphs presents difficult technical challenges, so it is simpler and more common to predict the \emph{linearized} graphs found in semantic parsing datasets using well-understood sequence models. The cost of this simplicity is that the predicted strings may not be well-formed graphs. We present recurrent neural network DAG grammars, a graph-aware sequence model that ensures only well-formed graphs while sidestepping many difficulties in graph prediction. We test our model on the Parallel Meaning Bank---a multilingual semantic graphbank. Our approach yields competitive results in English and establishes the first results for German, Italian and Dutch.
\end{abstract}

\section{Introduction}
\label{sec:introduction-1}

Semantic parsing is the task of mapping natural language to machine
interpretable meaning representations, which in turn can be expressed
in many different formalisms, including lambda calculus
\cite{Montague:1973}, dependency-based compositional semantics
\cite{liang2011learning}, frame semantics \cite{baker1998berkeley},
abstract meaning representations (AMR; \citealt{amr}), minimal
recursion semantics (MRS; \citealt{copestake2005minimal}), and
discourse representation theory (DRT; \citealt{kamp1981theory}).

\begin{figure}
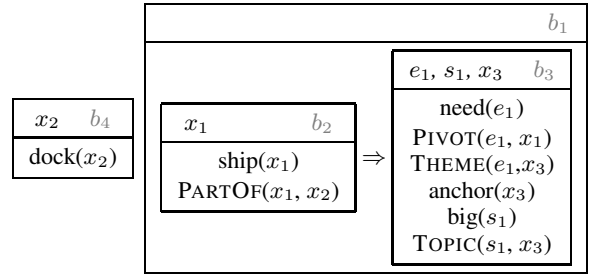
\footnotesize
\drs{$x_2$ \quad \textcolor{gray}{$b_4$}}{dock($x_2$)} 
\drs{\qquad\qquad\qquad\qquad\qquad\qquad\qquad\qquad   \textcolor{gray}{$b_1$}}{\drs{$x_1$ \qquad\qquad  \textcolor{gray}{$b_2$}}{ship($x_1$)\\\textsc{PartOf}($x_1$, $x_2$)} $\Rightarrow$ \drs{$e_1$, $s_1$, $x_3$ \quad \textcolor{gray}{$b_3$}}{need($e_1$)\\\textsc{Pivot}($e_1$, $x_1$)\\\textsc{Theme}($e_1$,$x_3$)\\anchor($x_3$)\\big($s_1$)\\\textsc{Topic}($s_1$, $x_3$)}}
\caption{The discourse representation structure for ``\textsl{Every
    ship in the dock needs a big anchor}''. For ease of reference in
  later figures, each box includes a variable corresponding to the box itself, at top right in \textcolor{gray}{gray}.}\label{fig:drs}
\end{figure}

\begin{figure}\footnotesize
    \centering
        \begin{tikzpicture}[
            level/.style={sibling distance=1cm,level distance=1.1cm},
            dot node/.style={draw,circle,minimum size=0.1cm,inner sep=0pt,outer sep=0pt,fill},
            edge from parent/.style={draw,-latex}
            ]

        \node[dot node,label={right:$b_1$/$\square$}] (b2) {}
            child {node [dot node,label={left:$b_2$/$\square$}] (b3) {} 
                child {node [xshift=-5mm,dot node, label={left:$x_1$/ship}] (x1) {}
                    child {node [xshift=-5mm,dot node, label={left:$x_2$/dock$^P$}] (x2) {}
                        edge from parent node[->,label={left:\textsc{PartOf}}] {}
                    }
                    edge from parent node[->,label={left:\textsc{Drs}}] {}
                }
                edge from parent node[->,label={left:\textsc{Imp}$_1$}] {}
            }
            child {node [dot node,label={right:$b_3$/$\square$}] (b4) {}
                child {node [xshift=5mm,dot node, label={right:$e_1$/need}] (e1) {}
                    child {node [xshift=5mm,dot node, label={left:$x_3$/anchor}] (x3) {}
                        child {node [xshift=15mm,yshift=11mm,dot node, label={right:$s_1$/big}] (s1) {}
                            edge from parent node[->,label={above:\textsc{TopicOf}}] {}
                        }
                        edge from parent node[->,label={left:\textsc{Theme}}] {}
                    }
                    edge from parent node[->,label={right:\textsc{Drs}}] {}
                }
                edge from parent node[->,label={right:\textsc{Imp}$_2$}] {}
            }
        ;
        
        \path (e1) edge[->,draw,-latex] node[label={above:\textsc{Pivot}}] {} (x1);
\end{tikzpicture}
    \caption{The DRS of Figure~\ref{fig:drs} expressed as a DAG.}
    \label{fig:graph}
\end{figure}
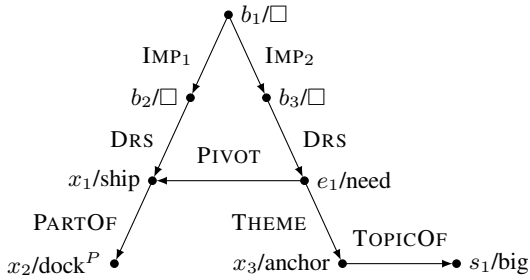

Explicitly or implicitly, a representation in any of these formalisms can be expressed as a directed acyclic graph (DAG). Consider the sentence ``\textsl{Every ship in the dock needs a big anchor}''. Its meaning representation, expressed as a Discourse Representation Structure (DRS, \citealt{kamp1981theory}), is shown in Figure~\ref{fig:drs}.\footnote{For simplicity, our examples do not show time representations, though these are consistently present in our data.} A DRS is drawn as a box with two parts: the top part lists variables for discourse referents (e.g.~$x_1$, $e_1$) and the bottom part can contain unary predicates expressing the type of a variable (e.g.~ship, need), binary predicates specifying relationships between variables (e.g.~\textsc{PartOf}, \textsc{Topic}), logical operators expressing relationships between nested boxes (e.g.~$\Rightarrow$,~$\lnot$), or binary discourse relations (e.g.,~\textsc{Result}, \textsc{Contrast}). To express a DRS as a graph, we represent each box as a node labeled $\square$; each variable as a node labeled by its associated unary predicate; and each binary predicate, logical operator, or discourse relation as an edge from the first argument to the second (Figure~\ref{fig:graph}). To fully realize the representation as a DAG, additional transformations are sometimes necessary: in DRS, when a box represents a \emph{presupposition}, as box $b_4$ does, the label of the node corresponding to the presupposed variable is marked (e.g.~$x_2$/dock$^P$); and edges can be reversed (e.g.~\textsc{Topic}($s_1$, $x_3$) becomes \textsc{TopicOf}($s_1$, $x_3$)).

Since meaning representations are graphs, semantic parsing should be
modeled as graph prediction. But how do we predict graphs? A popular
approach is to predict the \emph{linearized} graph---that is, the
\emph{string} representation of the graph found in most semantic
graphbanks. Figure~\ref{fig:penman} illustrates one style of
linearization using PENMAN notation, in which graphs are written as
well-bracketed strings which can also be interpreted as trees---note
the correspondence between the tree-like structure of
Figure~\ref{fig:graph} and the string in
Figure~\ref{fig:penman}.\footnote{Although PENMAN notation is now
  closely associated with AMR, it can represent quite arbitrary graphs
  as strings. Our actual implementation does not use PENMAN notation,
  but we use it here for expository purposes since it is relatively
  familiar; the underlying ideas are unchanged.} Each subtree is a
bracketed string starting with a node variable and its label
(e.g.~$b_2$/$\square$), followed by a list of \emph{relations}
corresponding to the outgoing edges of the node. A relation consists
of the edge label prefixed with a colon (:), followed by either the
subtree rooted at the target node (e.g.~:\textsc{Drs} ($x_1$/ship
:\textsc{PartOf}($x_2$/dock$^p$))), or a \emph{reference} to the
target node (e.g.~:\textsc{Pivot} $x_1$). By convention, if a node is
the target of multiple edges, then the leftmost one is written as a
subtree, and the remainder are written as references. Hence, every node
is written as a subtree exactly once.

The advantage of predicting linearized graphs is twofold. The first advantage is that graphbank datasets usually already contain linearizations, which can be used without additional work. These linearizations are provided by annotators or algorithms and are thus likely to be very consistent in ways that are beneficial to a learning algorithm. The second advantage is that we can use simple, well-understood sequence models \citep{gu2016incorporating,jia2016data,van2018exploring} to model them. But this simplicity comes with a cost: sequence models can predict strings that don't correspond to graphs---for example, strings with ill-formed bracketings or unbound variable names. While it is often possible to fix these strings with pre- or post-processing, we would prefer to model the problem in a way that does not require this.

Models that predict graphs are complex and far less well-understood than models that predict sequences. Fundamentally, this is because predicting graphs is difficult: every graph has many possible linearizations, so from a probabilistic perspective, the linearization is a latent variable that must be marginalized out \citep{li2018learning}. \citet{groschwitz2018amr} model graphs as trees, interpreted as the (latent) derivation trees of a graph grammar; \citet{lyu2018amr} model graphs with a conditional variant of the classic \citet{erdos+renyi} model, first predicting an alignment for each node of the output graph, and then predicting, for each pair of nodes, whether there is an edge between them. \citet{buys2017robust}, \citet{chen2018sequence}, and \citet{damonte2017incremental} all model graph generation as a sequence of actions, each aligned to a word in the conditioning sentence. Each of these models has a latent variable---a derivation tree or alignment---which must be accounted for via preprocessing or complex inference techniques.

\begin{figure}
\centering
\shortstack[l]{
($b_1$/$\square$\\
\quad :\textsc{Imp}$_1$($b_2$/$\square$\\
\quad\quad :\textsc{Drs}($x_1$/ship\\
\quad\quad\quad :\textsc{PartOf}($x_2$/dock$^p$)))\\
\quad :\textsc{Imp}$_2$($b_3$/$\square$\\
\quad\quad :\textsc{Drs}($e_1$/ need\\
\quad\quad\quad :\textsc{Pivot} $x_1$\\
\quad\quad\quad :\textsc{Theme}($x_3$/ anchor\\
\quad\quad\quad\quad :\textsc{TopicOf}($s_1$/ big)))))
}
\caption{\label{fig:penman} The DAG of Figure~\ref{fig:graph} expressed as a string.}
\end{figure}

Can we combine the simplicity of sequence prediction with the fidelity of graph prediction? We show that this is possible by developing a new model that predicts sequences through a simple string rewriting process, in which each rewrite corresponds to a well-defined graph fragment. Importantly, any well-formed string produced by our model has exactly one derivation, and thus no latent variables. We evaluate our model on the Parallel Meaning Bank (PMB, \citealt{abzianidze2017parallel}), a multilingual corpus of sentences paired with DRS representations. Our model performs competitively on English, and better than sequence models in German, Italian, and Dutch.

\section{Graph-aware string rewriting}\label{model:rewriting}

We use a grammar to model the process of rewriting. Formally, our grammar is a \emph{graph grammar}, specifically a \emph{restricted DAG grammar} \citep{bjorklund2016between}, a type of context-free graph grammar designed to model linearized DAGs. Since linearized DAGs are strings, we present it as a string-rewriting system, which can be described more compactly than a graph grammar while making the connection to sequences more explicit. The correspondence between string rewriting and graph grammars is given in the Appendix. 

A grammar in our model is defined by a set $\Sigma$ of
\textbf{terminal symbols} consisting of all symbols that can appear in
the final string---brackets, variable types, node labels, and edge
labels; a set $N$ of $n+1$ \textbf{nonterminal symbols} denoted by
$\{L, T_0, \dots, T_n\}$, for some maximum value~$n$; an unbounded set
$V$ of \textbf{variable references} $\{\$1, \$2, \dots \}$; and a set
of productions, which are defined below.

We say that $T_0$ is the \textbf{start symbol}, and for each symbol
$T_i \in N$, we say that $i$ is its \textbf{rank}, and we say that $L$
has a rank of 0. A nonterminal of rank $i$ can be written as a
\textbf{function} of $i$ variable references---for example, we can
write $T_2(\$1, \$2)$. By convention, we write the rank-0 nonterminals
$L$ and $T_0$ without brackets. Productions in our grammar take the
form $\alpha \rightarrow \beta$, where $\alpha$ is a function of rank
$i$ over $\$1,\dots, \$i$; and $\beta$ is a linearized graph in PENMAN
format, with each of its subtrees replaced by either a function or a
variable reference. Optionally, the variable name in $\beta$ may
replace one of the variable references in $\alpha$. All variable
references in a production must appear at least twice. Hence every
variable reference in $\alpha$ must appear at least once in $\beta$,
and variables that do not appear in $\alpha$ must appear at least
twice in $\beta$.

To illustrate, we will use the following grammar, which can generate the string in Figure~\ref{fig:penman}, assuming $L$ can also rewrite as any node label.\begin{align*}
    T_0 &\rightarrow (b/\square \textsc{~:Imp}_1~T_1(\$1) \textsc{~:Imp}_2~T_1(\$1)) \tag{$r_1$}\\
    T_0 &\rightarrow (x/L) \tag{$r_2$}\\
    T_0 &\rightarrow (s/L) \tag{$r_3$}\\
    T_0 &\rightarrow (x/L \textsc{~:TopicOf~} T_0) \tag{$r_4$}
\end{align*}\vspace{-9mm}
\begin{align*}
    T_1(\$1) &\rightarrow (b/\square \textsc{~:Drs~} T_1(\$1)) \tag{$r_5$}\\
    T_1(\$1) &\rightarrow (e/L \textsc{~:Pivot~} \$1 \textsc{~:Theme~} T_0) \tag{$r_6$}\\
    T_1(x) &\rightarrow (x/L \textsc{~:PartOf~} T_0) \tag{$r_7$}
\end{align*}

Our grammar derives strings by first rewriting the start symbol $T_0$, and at each subsequent step rewriting the leftmost function in the partially derived string, with special handling for variable references described below. A derivation is complete when no functions remain. 

We illustrate the rewriting process in Figure~\ref{fig:derivation}. The start symbol $T_0$ at step 1 is rewritten by production $r_1$ in step 2, and the new $b$ variable introduced at this step is deterministically renamed to the unique name $b_1$. In step 3, the leftmost $T_1(\$1)$ is rewritten by production $r_5$, and the new $b$ variable is likewise renamed to the unique $b_2$. All productions apply in this way, simply replacing a left-hand-side function with a right-hand side expression. These rewrites are
coupled with a mechanism to correctly handle multiple references to shared variables, as illustrated in Step 4 when production $r_7$ is applied. In this production, the left-hand-side function applies to the $x$ variable naming the right-hand-side node. When this production applies, $x$ is renamed to the unique $x_1$ as in previous steps, but because it appears in the left-hand-side, the reference $\$1$ is bound to this new variable name throughout the partially-derived string. In this way, the reference to $x_1$ is passed through the subsequent rewrites, becoming the target of a \textsc{Pivot} edge at step 10.
Derivations of a DAG grammar are context-free (Figure~\ref{fig:dertree}).\footnote{However, the language of derived strings may not be context-free due to variable references.}

Our model requires an explicit grammar like the one in $r_1...r_7$, which we obtain by converting each DAG in the training data into a sequence of productions. The conversion yields a single, unique sequence of productions via a simple linear-time algorithm that recursively decomposes a DAG into subgraphs \citep{bjorklund2016between}. Each subgraph consists of single node and its outgoing edges, as exemplified by the PENMAN-formatted right-hand-sides of $r_1$ through $r_7$.
Each outgoing edge points to a nonterminal symbol representing a subgraph. If a subgraph does not share any nodes with its siblings, it is represented by $T_0$. But if any subgraphs share a node, then a variable reference must refer for this node in the production associated with the lowest common ancestor of all its incoming edges. For example, in
Figure~\ref{fig:penman}, the common ancestor of the two edges targeting
$x_1$ is the node $b_1$, so production $r_1$ must contain two copies of variable 
reference \$1 to account for this. A more mathematical account
can be found along with a proof of correctness in \citet{bjorklund2016between} 
Our implementation follows their description unchanged.

\begin{figure*}
\begin{align*}
\begin{array}{rlll}
\textrm{Step} & \textrm{Action} & \textrm{Production} & \textrm{Result} \\ \hline
  1 & \textrm{START} & \textrm{start} &\underline{T_0} \\
  2 & \textrm{GEN-FRAG} & r_1  &\textcolor{blue}{(b_1/\square \textsc{~:Imp}_1~\underline{T_1(\$1)} \textsc{~:Imp}_2~T_1(\$1))} \\
  3 & \textrm{GEN-FRAG} & r_5  &(b_1/\square \textsc{~:Imp}_1~\textcolor{blue}{(b_2/\square \textsc{~:Drs~} \underline{T_1(\$1))}} \textsc{~:Imp}_2~T_1(\$1)) \\
  4 & \textrm{GEN-FRAG} & r_7  &(b_1/\square \textsc{~:Imp}_1~(b_2/\square \textsc{~:Drs~} \textcolor{blue}{(x_1/\underline{L} \textsc{~:PartOf~} T_0)} \textsc{~:Imp}_2~T_1(x_1)) \\
  5 & \textrm{GEN-LABEL} & L\rightarrow\textrm{ship}  &(b_1/\square \textsc{~:Imp}_1~(b_2/\square \textsc{~:Drs~} (x_1/\textcolor{blue}{\textrm{ship}} \textsc{~:PartOf~} \underline{T_0}) \textsc{~:Imp}_2~T_1(x_1)) \\
  6 & \textrm{GEN-FRAG} & r_2  &(b_1/\square \textsc{~:Imp}_1~(b_2/\square \textsc{~:Drs~} (x_1/\textrm{ship} \textsc{~:PartOf~} \textcolor{blue}{(x_2/\underline{L})}) \\
  & & & \textsc{~:Imp}_2~T_1(x_1)) \\
  7 & \textrm{GEN-LABEL} & L\rightarrow\textrm{dock}^p &(b_1/\square \textsc{~:Imp}_1~(b_2/\square \textsc{~:Drs~} (x_1/\textrm{ship} \textsc{~:PartOf~} (x_2/\textcolor{blue}{\textrm{dock}^p}))\\
  & & & \textsc{~:Imp}_2~\underline{T_1(x_1)}) \\
  8 & \textrm{REDUCE} & -- & =\\
  9 & \textrm{REDUCE} & -- & =\\
  8 & \textrm{GEN-FRAG} & r_5 &(b_1/\square \textsc{~:Imp}_1~(b_2/\square \textsc{~:Drs~} (x_1/\textrm{ship} \textsc{~:PartOf~} (x_2/\textrm{dock}^p))\\  
    & &  &\qquad~~\textsc{~:Imp}_2~\textcolor{blue}{(b_3/\square \textsc{~:Drs~} \underline{T_1(x_1)})}) \\
  9 & \textrm{GEN-FRAG} & r_6 &(b_1/\square \textsc{~:Imp}_1~(b_2/\square \textsc{~:Drs~} (x_1/\textrm{ship} \textsc{~:PartOf~} (x_2/\textrm{dock}^p))\\  
    & &  &\qquad~~\textsc{~:Imp}_2~(b_3/\square \textsc{~:Drs~} \textcolor{blue}{(e_1/\underline{L} \textsc{~:Pivot~} x_1 \textsc{~:Theme~} T_0))}) \\
  10 & \textrm{GEN-LABEL} & L\rightarrow\textrm{need} &(b_1/\square \textsc{~:Imp}_1~(b_2/\square \textsc{~:Drs~} (x_1/\textrm{ship} \textsc{~:PartOf~} (x_2/\textrm{dock}^p))\\  
    & &  &\qquad~~\textsc{~:Imp}_2~(b_3/\square \textsc{~:Drs~} (e_1/\textcolor{blue}{\textrm{need}} \textsc{~:Pivot~} x_1 \textsc{~:Theme~} \underline{T_0})))
\end{array}
\end{align*}
\caption{A partial derivation of the string in Figure~\ref{fig:penman}. The stack operations follow closely each step in the derivation, where GEN-FRAG and GEN-LABEL are invoked when rewriting a non-terminal $T$ and a terminal $L$ respectively. In the result of each step, the leftmost function is underlined, and is rewritten in the fragment in blue in the next step. On the other hand, a REDUCE operation is invoked when a generated fragment does not contain non-terminals $T$ to expand further (in this partial derivation, this is the case of the result of production $r_2$).}
\label{fig:derivation}
\label{table:derivation}
\end{figure*}

\section{Neural Network Realizer}\label{model:nnr}

We model graph parsing with an encoder-decoder architecture that
takes as input a sentence $w$ and outputs a directed acyclic graph
$G$ derived using the rewriting system of
Section~\ref{model:rewriting}. Specifically, we model its derivation tree in top-down, left-to-right order as a sequence of actions $a = a_1 \dots a_{|a|}$, inspired by Recurrent Neural Network Grammars \cite[RNNG,][]{dyer2016recurrent}. As in RNNG, we use a stack to store partial derivations.

We model two types of actions: \textbf{GEN-FRAG} rewrites $T_i$
nonterminals, while \textbf{GEN-LABEL} rewrites
$L$ nonterminals, always resulting in a leaf of the derivation tree. A
third \textbf{REDUCE} action is applied whenever a subtree of the derivation tree is complete, and since the number of subtrees is known in advance, it is applied deterministically. For example, 
when we predict $r_1$, this determines that we
must rewrite an $L$ and then recursively rewrite two copies of $T_1(\$1)$ and then
apply \textbf{REDUCE}.
Hence graph generation reduces to predicting rewrites only.

We define the probability of generating graph $G$ conditioned of input sentence $w$ as follows:
\begin{equation}
 p(G|w) = p(a|w) = \prod_{i=1}^{|a|} p(a_i|a_{<i},w)
\end{equation} 

\paragraph{Input Encoder} We represent the $i$th word $w_i$ of input sentence
$w=w_1 \dots w_{|w|}$ using both learned and
pre-trained word embeddings ($\mathbf{w}_i$ and $\mathbf{w}_i^p$ respectively),
lemma embedding ($\mathbf{l}_i$), part-of-speech embedding ($\mathbf{p}_i$), 
universal semantic tag \cite{abzianidze2017towards} embedding ($\mathbf{u}_i$), and dependency label embedding ($\mathbf{d}_i$).\footnote{Universal semantic tags are language neutral tags intended to characterize lexical semantics.} An input $\mathbf{x}_i$ is computed as the weighted concatenation of these features followed by a non-linear
projection (with vectors and matrices in {\bf bold}):
\begin{equation}
\mathbf{x}_i = \tanh(\mathbf{W^{(1)}} [{\bf w}_i ; {\bf w}_i^p ; {\bf l}_i ; {\bf p}_i ; {\bf u}_i ; {\bf d}_i])
\end{equation}
Input $\mathbf{x}_i$ is then encoded with a bidirectional LSTM, yielding contextual representation $\mathbf{h}_i^e$.
%\begin{equation}
%\hspace*{-.2cm}\mathbf{h}_t = \text{LSTM}(\mathbf{x}_t, \mathbf{\overrightarrow{{h}}}_{t-1}) ; \text{LSTM}(\mathbf{x}_t, \mathbf{\overleftarrow{{h}}}_{t+1})
%\end{equation}

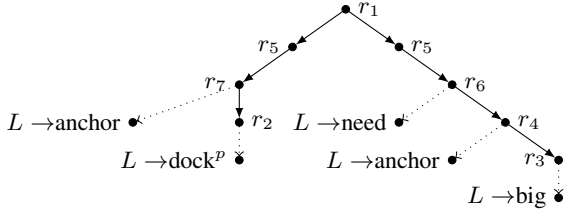
\begin{figure}[t]\small
\centering
        \begin{tikzpicture}[
            level/.style={sibling distance=1.4cm,level distance=5mm},
            dot node/.style={draw,circle,minimum size=0.1cm,inner sep=0pt,outer sep=0pt,fill},
            edge from parent/.style={draw,-latex}
%             edge2 from parent/.style={draw,-latex}
            ]

        \node[dot node,label={right:$r_1$}] (r1) {}
            child {node [dot node,label={left:$r_5$}] (r5) {} 
                child {node [xshift=-7mm,dot node, label={left:$r_7$}] (r7) {}
                    child {node [xshift=-7mm,dot node, label={left:$L\rightarrow$anchor}] (lr7) {}
			  (r7) edge[->, dotted] (lr7);
		    }
                    child {node [xshift=-7mm,dot node, label={right:$r_2$}] (r2) {}
			child {node [dot node, label={left:$L\rightarrow$dock$^p$}] (lr2) {}
			  (r2) edge[->, dotted] (lr2);
		      }
                    }
                }
            }
            child {node [dot node,label={right:$r_5$}] (r5p) {}
                child {node [xshift=7mm,dot node, label={right:$r_6$}] (r6) {}
                    child {node [dot node, label={left:$L\rightarrow$need}] (lr6) {}
		      (r6) edge[->, dotted] (lr6);
                    }
		    child {node [dot node, label={right:$r_4$}] (r4) {}
			child {node [dot node, label={left:$L\rightarrow$anchor}] (lr4) {}
			  (r4) edge[->, dotted] (lr4);
			}
			child {node [dot node, label={left:$r_3$}] (r3) {}
			  child {node [dot node, label={left:$L\rightarrow$big}] (lr3) {}
			    (r3) edge[->, dotted] (lr3);
			  }
			}
		      }
                    }
                };
        
%         \path (e1) edge[->,draw,-latex] node[label={above:\textsc{Pivot}}] {} (x1);
\end{tikzpicture}
    \caption{A derivation tree corresponding to Figure~\ref{table:derivation}. Solid edges rewrite $T_n$ nonterminals, while dotted rewrite $L$ nonterminals.}
    \label{fig:dertree}
\end{figure}

% In a standard RNNG model prediction consists of two steps, each modeled by a dedicated LSTM: first, an action is predicted via an actionLSTM and subsequently a stackLSTM predicts either a terminal or non-terminal conditioned on the action selected. However, the fragment in our grammar already encodes information about how many children graph to generate and how this connect together, as given by the arguments of the edge labels. For this reason, our model is fully \textit{determistic} with respect to the actions.

\paragraph{Graph decoder} Since we know in advance whether the next action is \textbf{GEN-FRAG} or \textbf{GEN-LABEL}, we use different models for them.

\textbf{GEN-FRAG}. If step $t$ rewrites a $T$ nonterminal, we predict the production $y_t$ that rewrites it
using context vector~$\mathbf{c}_t$ and incoming edge
embedding~$\mathbf{e}_t$. To obtain $\mathbf{c}_t$ we use \textit{soft
  attention} \cite{luong2015effective} and weight each input hidden
representation $\mathbf{h}^{e}_i$ to decoding hidden
state~$\mathbf{h}_t^d$:%. The computation for target fragment $y_t$ is as follows:
\begin{equation}
\label{itemgenerator}
\begin{aligned}
  \textrm{score}(\mathbf{h}_i^e, \mathbf{h}_t^d) &= \mathbf{h}_i^e \mathbf{W}^{(2)} \mathbf{h}_t^d\\
  \alpha_{ti} &= \frac{\exp(\textrm{score}(\mathbf{h}_i^e, \mathbf{h}_t^d))}{\sum_{i'} \exp(\textrm{score}(\mathbf{h}_{i'}^e, \mathbf{h}_t^d))}\\
  \mathbf{c}_t &= \sum_{i=1}^n \alpha_{ti} \mathbf{h}_i^e\\
  \mathbf{y}_t &= \mathbf{W}^{(3)} \mathbf{c}_t + \mathbf{W}^{(4)} \mathbf{e}
\end{aligned}
\end{equation}
The contribution of~$\mathbf{c}$ and~$\mathbf{e}$ is weighted by matrices~$\mathbf{W}^{(3)}$ and~$\mathbf{W}^{(4)}$, respectively.

We then update the stackLSTM representation using the embedding of the
non-terminal fragment~$\mathbf{y}_t$ (denoted as $\mathbf{y}_{t}^{e}$), as follows:
\begin{equation}
    \label{lstmupdate}
    \mathbf{h}^{d}_{t+1} = \text{LSTM}(\mathbf{y}_{t}^{e}, \mathbf{h}^{d}_t)
\end{equation}

\textbf{GEN-LABEL}. Labels $L$ can be rewritten to either semantic constants (e.g.,~`speaker', `now',
`hearer') or unary predicates that often corresponds to the lemmas of the input words (e.g.,~`love')
or. We predict the former using a model identical to the 
one for GEN-FRAG. For the latter, we use a \textit{selection mechanism} to
choose an input lemma to copy to output. We model selection
following \citet{liu2018discourse}, assigning each
input lemma a score $o_{ji}$ that we then pass
through a softmax layer to obtain a distibution:

\begin{equation}
  \begin{aligned}
   o_{ji} &=\mathbf{h}_j^{dT}\mathbf{W}^{(5)}\mathbf{h}^{e}_{i}\\
   p_{i}^{copy} &= \textsc{SoftMax}(o_{ji})
  \end{aligned}
\end{equation}
where $\mathbf{h}_{i}$ is the encoder hidden state for word $w_i$.

We allow the model to learn whether to use soft-attention or the
selection mechanism through a binary classifier, conditioned on the
decoder hidden state at time $t$, $\mathbf{h}^{d}_t$. Similar to Equation~\eqref{lstmupdate}, we update the
stackLSTM with the embedding of terminal predicted.\footnote{In the PMB, each terminal is annotated for sense (e.g.~`n.01', `s.01') and presupposition (e.g.~for `dock$^{p}$' in Figure~\ref{fig:penman}) as well. We predict both the sense tag and whether a terminal is presupposed or not independently conditioned on the current stackLSTM state and the embedding of the main terminal labels but are \textbf{not} used to update the state of the stackLSTM.}

\textbf{REDUCE}. When a reduce action is applied, we
use an LSTM to compose the fragments on top of the stack. Using the
derivation tree in Figure~\ref{fig:dertree} as reference, let
[$\mathbf{c}_1,...,\mathbf{c}_n$] denote the embeddings of one or more sister nodes $r_i$ and $p_u$ the embedding of their parent node, which we refer to as \textit{children} and \textit{parent fragments} respectively. A
reduce operation runs an LSTM over the children fragments and the
parent fragment in order and then uses the final state~$u$ to update
the stack LSTM as follows:
\begin{equation}
\begin{aligned}
\mathbf{i} &= [\mathbf{c}_1 ...\mathbf{c}_n, \mathbf{p}_u]\nonumber\\
\overrightarrow{\mathbf{u}} &= LSTM(\mathbf{i}_t, \mathbf{h}_t^c)\nonumber\\
\mathbf{h}^{d}_{t+1} &= LSTM(\mathbf{u}, \mathbf{h}^{d}_t)
\end{aligned}
\end{equation}

The models are trained to minimize a cross-entropy loss objective $J$ over the sequence of gold actions $a_i$ in the derivation:
\begin{equation}
J = - \sum^{|a|}_{i=1}log \; p(a_i)
\end{equation}

\section{Experimental Setup}
\label{sec:experimental-setup}

%\section{Data}
We evaluated our model on the Parallel Meaning Bank (PMB;
\citealt{abzianidze2017parallel}), a semantic bank where sentences in
English, Italian, German, and Dutch have been annotated following
Discourse Representation Theory \citep{kamp2013discourse}. Lexical
predicates in PMB are in English, even for non-English languages.
Since this is not compatible with our copy mechanism, we revert
predicates to their orignal language by substituting them with the
lemmas of the tokens they are aligned to. In our experiments we used
both PMB v.2.1.0 and v.2.2.0\footnote{English data is available for
  both releases at \url{https://github.com/RikVN/DRS_parsing}; for the
  other languages, we used the officially released data available at
  \url{http://pmb.let.rug.nl/data.php}}; we included the former
release in order to compare against the state-of-the-art seq2seq
system of \citet{van2018exploring}. Statistics on the data and the
grammar extracted from v.2.2.0 are reported in Table~\ref{tab:data}.

\begin{table}[]
    \centering
    \scalebox{0.8}{
        \begin{tabular}{lrrcccr} \thickhline
%             & \multicolumn{5}{c}{gold} & \multicolumn{3}{c}{silver}\\
          & \multicolumn{3}{c}{train} & dev & test \\
          \textbf{Gold}   & \#inst. & \#frags & avg. rank & \#inst. & \#inst.  \\\thickhline

             en & 4,574 & 1,105 & 1.64 & 682 & 650\\ 
             it & --- & --- & --- & 387 & 386 \\
             de & --- & --- & --- & 736 & 735 \\
             nl & --- & --- & --- & 356 & 355  \\\thickhline
\multicolumn{3}{}{}\\\thickhline
&  \multicolumn{3}{c}{train}  \\
\textbf{Silver} &  \#inst. & \#frags & avg. rank \\\thickhline
en       & 52,120 & 11,206 & 1.90\\  
it       & 2,546 & 1,051 & 1.52\\
de & 3,274 & 1,781& 1.54\\
nl &  874 & 789 & 1.45\\\thickhline
        \end{tabular}
    }
    \caption{Data statistics of PMB v.2.2.0.}
    \label{tab:data}
\end{table}

\subsection{Converting DRSs to Graphs}
\label{sec:conv-drss-graphs}

In this section we discuss how DRSs are converted to \textit{acyclic}, \textit{single-rooted}, and \textit{fully-instantiated} graphs (i.e.,~how to translate Figure~\ref{fig:drs} to Figure~\ref{fig:graph}). In general, box structures are converted
to acyclic graphs by rendering boxes and lexical predicates as nodes, 
while conditions, operators, and discourse relations become edge
labels between these nodes.

We consider \textit{main} boxes (see $b_2$, $b_3$, and $b_4$ in
Figure~\ref{fig:drs} separately from \textit{presuppositional} boxes
(see~$b_1$), which represent instances that are presupposed in the
wider discourse context (e.g.,~definite expressions). Using
Figure~\ref{fig:graph} as an example, $b_2$, $b_3$ and $b_4$ become nodes in the graph and material implication (IMP stands for
$\Rightarrow$) becomes an edge label. If an operator or a relation is
binary, as in this case, we number the edge label so as to preserve
the order of the operands.

For each node in a \textit{main} box, we expand the graph by adding
all relations and variables belonging to it. We identify the head of
the first relation or the first referent mentioned as the \textit{head variable}. These are `ship ($x_1$)' for~$b_2$, and `need($e_1$)'
for~$b_3$. We attach the head variable as a child of the box-node and
follow the relations recursively to expand the subgraph. If
while expanding a graph a variable in a condition is part of a
presuppositional box, we introduce it as a new node and add to its
label the superscript~$^p$. When expanding the DAG along the edge
{PartOf}, since $x_2$ is also in the presuppositional box in
Figure~\ref{fig:drs}, we attach the node `dock$^p$'.  Graphs
extracted this way are mostly acyclic except for adjectival phrases
and relative clauses where state variables can be themselves root
(e.g., big ``a.01'' $s_1$). We get rid of these extra roots by
reversing the direction of the edge involved and adding an `-of' to
the edge label to flag this change (see \textsc{Topic-of}).

\subsection{System Comparison}
\label{sec:systems-experiments}

We compared the performance of our graph parser ({seq2graph} below)
with a sequence-to-sequence model (enchanced with a copy mechanism)
which decodes to a string linearization of the graph similar to the one shown in Figure~\ref{fig:penman} ({seq2seq + copy} below). We also
compare against the recently proposed model of
\citet{van2018exploring}; they introduce a seq2seq model that
generates a DRS as a concatenation of \textit{clauses}, essentially a
flat version of the standard box notation. The decoded string is made
aware of the overall graph structure during preprocessing where
variables are replaced by indices indicating when they were first
introduced and their recency. In contrast, we model the graph
structure explicitly. \citet{van2018exploring} experimented with both
word and character-based models, as well as with an ensemble of both,
using word embedding features. Since all our models are word-based, we
compare our results with their best word model, using word embedding
features only (trained using 10-fold cross validation).

\subsection{Model Configurations}

In addition to word embeddings\footnote{We used word embeddings pre-trained with Glove and available at \url{https://nlp.stanford.edu/projects/glove/}.}, we also report on
experiments which make use of additional features. Specifically, for
each word we add information about its universal PoS tag, lemma,
universal semantic tag, and dependency label.\footnote{Universal PoS
  tags, lemmas and dependency labels for all languages were obtained
  using pretrained UDPipe models available at
  \url{http://ufal.mff.cuni.cz/udpipe\#download}; gold-standard
  universal semantic tags were extracted from the PMB release.}

In Section~\ref{model:rewriting}, we mentioned that given a production $\alpha \rightarrow \beta$, variable references in $\alpha$ should appear at least once in $\beta$ (i.e.,~they should have the same rank). In all experiments so far, we did not model this constraint explicitly to investigate whether the model is able by default to predict rank correctly. However, in exploring model configurations we also report on whether adding this constrant leads to better performance . %(\textit{+constraint}).

\subsection{Cross-lingual Experiments}

We conducted two sets of experiments: one monolingual ({mono} below)
where we train and test on the same language and one cross-lingual
({cross} below), where we train a model on English and test it on one
of the other three languages. The goal of our cross-lingual
experiments is to examine whether we need data in a target language at
all since the semantic representation itself is language agnostic and
lexical predicates are dealt with via the copy mechanism. Most of the
features mentioned above are cross-linguistic and therefore fit both
mono and cross-lingual settings, with the exception of lemma and word
embeddings, where we exclude the former and replaced the latter with
multilingual word embeddings.\footnote{We experimented with embeddings
  obtained with iterative procrustes (available at
  \url{https://github.com/facebookresearch/MUSE}) and with Guo et
  al. \shortcite{guo2016representation}'s `robust projection' method
  where the embedding of non-English words is computed as the weighted
  average of English ones. We found the first method to perform better
  on cross-lingual word similarity tasks and used it in our
  experiments.}
  
\subsection{System settings}
For training, we used the Adam optimizer \cite{kingma2014adam} with an initial learning rate of 0.001 and a decay rate of 0.1 every 10 epochs. Randomly initialized and pre-trained word embeddings have a dimensionality of 128 and 100 respectively, and all other features a dimensionality of 50. In all cross-lingual experiments, the pre-trained word embeddings have a dimensionality of 300. The LSTMs in the encoder and the decoder have a dimensionality of 150 and non-terminal and terminal embeddings during decoding have a dimensionality of 50. The system is trained for 30 epochs, with the best system chosen based on dev set performance.

\subsection{Evaluation Metric} \label{evaluation}

We evaluated our system by scoring the similarity between predicted
and gold graphs. We used Counter \citep{van2018exploring}, an
adaptation of Smatch \citep{cai2013smatch} to Discourse Representation
Structures where graphs are first transformed into a set of `source
node -- edge label -- target node' triples and the best mapping
between the variables is found through an iterative hill-climbing
strategy. Furthermore, Counter checks whether DRSs are well-formed in
that all boxes should be connected, acyclic, with fully instantiated
variables, and correctly assigned sense tags.

It is worth mentioning that there can be cases where our parser generates ill-formed graphs according to Counter; this is however not due to the model itself but to the way the graph is converted back in a format accepted by Counter.

All results shown are an averages over 5~runs.

\section{Results}
\label{sec:results}

\paragraph{System comparison}
Table~\ref{baseline_comp} summarizes our results on the PMB gold data
(v.2.1.0, test set). We compare our graph decoder against the system
of \citet{van2018exploring} and our implementation of a seq2seq model,
enhanced with a copy mechanism. Overall, we see that our graph decoder
outperforms both models. Moreover, it reduces the number of illformed
representations without any specific constraints or post-processing in
order to ensure the well-formedness of the semantics of the output.

The PMB (v.2.1.0) contains a large
number of silver standard annotations which have been only partially
manually corrected (see Table~\ref{tab:data}). Following
\citet{van2018exploring}, we also trained our parser on both silver
and gold standard data combined. As shown in Table~\ref{silveren},
increasing the training data improves performance but the difference is not as dramatic as in \citet{van2018exploring}. 
We found that this is because our parser requires graphs that are fully instantiated -- all unary predicates (e.g.~\textit{ship(x)}) need to be present for the graph to be fully connected, which is often not the case for silver graphs. Our model is at a disadvantage since it could exploit less
training data; during grammar extraction we could not process around 20K sentences and in some cases could not reconstruct the whole graph,
as shown by the \textit{conversion score}.\footnote{The conversion score is computed using Counter where we consider the converted graph representation as a `predicted' DRS and the original DRS structure as gold data. As a comparison, converting the \textit{gold} DRS sturctures yields a conversion score of 99.8\%.}

\begin{table}[t]
\begin{center}
\begin{small}
%\scalebox{0.72}{
\begin{tabular}{@{}l@{~~}cccr@{}} \thickhline
 & P & R & F1 & illformed\\
\thickhline
 \citet{van2018exploring} & --- & --- & 72.80 & 2.0\%\\
%\cline{2-6}
 seq2seq + copy & 75.57 & 67.27 & 71.18 & 4.12\%\\
%\hline
 seq2graph & 75.51 & 71.69 & {\bf 73.55} & {\bf 0.40}\%\\\thickhline
\end{tabular}
\end{small}
%}
\caption{Model performance (Precision, Recall, F$_1$) on PMB data
  (v.2.1.0, test set); models were trained on \emph{gold} standard data.}
\label{baseline_comp}
\end{center}
\end{table}

\begin{table}[t]
\begin{center}
\begin{small}
%\scalebox{0.72}{
\begin{tabular}{@{}l@{~~~}c@{~~}c@{~}c@{~~}c@{~~}c@{}} \thickhline
& \#sents & \shortstack{conversion\\score} & F1 & illformed\\
\thickhline
\citet{van2018exploring} & 73,778 & 100\% & 82.7 & 1.10\%\\
seq2seq+copy & 56.694 & 89.58\% & 74.67 & 13.45\%\\
seq2graph & 56.694 & 89.58\% & 77.1 & 0.90\%\\ \thickhline
\end{tabular}
\end{small}
\caption{Model performance (Precision, Recall, F$_1$) on PMB data
  (v.2.1.0, test set); models were trained on \emph{gold} and
  \emph{silver} standard data combined.}
%}
\label{silveren}
\end{center}
\end{table}

\paragraph{Model configurations}
Table~\ref{Baseline} reports on various ablation experiments investigating which features and combinations thereof perform best. The experiments were conducted on the
development set (PMB v2.2.0). We show a basic version of our seq2graph
model with word embeddings, to which we add information about rank (+restrict). We also experimented with the
full gamut of additional features (+feats) as well as with ablations
of individual feature classes. For comparsion, we also show the
performance of a graph-to-string model (seq2seq+copy).

As can be seen, all linguistic features seem to improve
performance. Restricting fragment selection by rank does not seem to improve the overall result showing that our baseline model is already able to predict fragments with the correct rank throughout the derivation. Subsequent experiments report results with this model using all linguistic features, unless otherwise specified.

\begin{table}[t]
\begin{center}
%\scalebox{0.72}{
\begin{small}
\begin{tabular}{@{~}l@{~~~}c@{~~}c@{~~}c@{~~}c@{~~}} \thickhline
& P & R & F1 & illformed\\ 
\thickhline
seq2seq + copy & 60.29 & 74.09 & 66.48 & 10.60\%\\
\hline
seq2graph & \textbf{70.69} & \textbf{74.46} & \textbf{72.53} & 0.10\%\\
+restrict & 70.50 & 74.64 & 72.51 & 0.70\%\\
\hline
\textit{+feats} & \textbf{72.51} & \textbf{76.44} & \textbf{74.42} & 0.60\%\\
-lemmas & 71.53 & 76.28 & 73.83 & 0.32\%\\
-pos & 72.11 & 75.99 & 74.00 & 0.35\%\\
-semtag & 70.21 & 74.45 & 72.27 & 0.53\%\\
-words & 70.45 & 74.59 & 72.46 & 0.64\%\\
-dep & 72.16 & 76.40 & 74.22 & 0.50\%\\\hline
\end{tabular}
%}
\end{small}
\caption{Model performance (Precision, Recall, F$_1$) on PMB
  (v2.2.0, development set).}
\label{Baseline}
\end{center}
\end{table}

\paragraph{Cross-lingual experiments} Results on DRT parsing
for languages other than English are reported in
Table~\ref{monoother}.  There is no gold standard training data for
non-English languages in the PMB (v.2.2.0). We therefore trained our
parser on silver standard data but did use the provided gold standard
data for development and testing (see Table~\ref{tab:data}).  We
present two versions of our parser, one where we train and test on the
same language (s2g mono-silver) and another one where a model is
trained on English but tested on the other languages (s2g
cross-silver). We also show the results of a sequence-to-sequence
model enhanced with a copy mechanism.

In the monolingual setting, our graph parser outperforms the seq2seq
baseline by a large margin; we hypothesize this is due to the large
percentage of ill-formed semantics, mostly due to training on silver
data. The difference in performance between our cross-lingual parser and the
monolingual parser for all languages is small, and in Dutch the two parsers perform on par, suggesting that English data
and language independent features can be leveraged to build parsers in
other languages when data is scarse or event absent. We also conducted
various ablation studies to examine the contribution of individual
features to cross-linguistic semantic parsing. Our experiments
revealed that universal semantic tags are most useful, while
the multilingual word embeddings that we have tested with are not. We refer the interested reader to the supplementary material for more detail on these experiments.

\begin{table}[t]
\begin{center}
%\scalebox{0.7}{
\begin{small}  

  \begin{tabular}{@{}lcccr@{}}         \thickhline
\multicolumn{1}{c}{\emph{Italian}}           & P & R & F1 & ill \\         \thickhline
s2s + copy mono-silver & 61.33 & 72.42 & 66.41 & 14.80 \\
s2g mono-silver & 74.50 & 77.27 & 75.86 & 0.10 \\
s2g cross-silver & 70.35 & 71.91 & 71.12 & 0.00 \\\thickhline
\multicolumn{4}{c}{} \\\thickhline
\multicolumn{1}{c}{\emph{German}}           & P & R & F1 & ill \\         \thickhline
s2s + copy mono-silver& 56.86 & 65.40 & 60.83 & 10.67 \\
s2g mono-silver& 66.44 & 69.34 & 67.86 & 0.80 \\
s2g cross-silver& 66.14 & 65.72 & 63.50 & 0.40 \\\thickhline
\multicolumn{4}{c}{} \\\thickhline
\multicolumn{1}{c}{\emph{Dutch}}           & P & R & F1 & ill \\         \thickhline
s2s + copy mono-silver& 54.27 & 64.17 & 58.81 & 10.67\\
s2g mono-silver& 63.50 & 68.37 & 65.84 & 0.86\\
s2g cross-silver& 62.94 & 67.32 & 65.06 & 0.50\\\thickhline
    \end{tabular}
\end{small}
 %   }
\end{center}
\caption{Model performance across languages (Precision, Recall,
  F$_1$). Results are reported on the PMB test set (v.2.2.0) for each language.}
\label{monoother}
\end{table}

\paragraph{Error Analysis}
We further analyzed the output of our parser to gain insight as to
what parts of meaning representation are still challenging.
Table~\ref{pmbeval} shows a more detailed break-down of system output
as computed by Counter, where \emph{operators} (e.g., negation,
implication), roles (i.e.,~binary relations, such as `Theme'), concepts (i.e.,~unary predicates like `ship'), and
\emph{synsets} (i.e.,~sense tags like `n.01') are scored
separately.  Synsets are further broken down into into `Nouns',
`Verbs', 'Adverbs', and `Adjectives'.  We compare our best seq2graph models (\textit{+feats}) trained on the PMB v.2.2.0, gold and gold+silver data respectively.

Adding silver data helps with semantic elements (operators, roles and concepts), but does not in the case of sense prediction where the only category that benefits from additional data are is nouns. We also found that ignoring the prediction of sense tags altogether  helps with the performance of both models.

\begin{table}[t]
    \centering
\scalebox{0.9}{
\begin{small}
    \begin{tabular}{lccc} \thickhline
         & seq2graph(gold) & seq2graph(silver) \\ \thickhline 
%         \hline
         all clauses & 74.42 & \textbf{76.36}\\       
%         \hline
         DRS operators & 88.05 & \textbf{89.36} \\
	 Roles & 72.95 & \textbf{74.03} \\
         Concepts & 71.13 & \textbf{74.42} \\
         \thickhline
         Synsets ~~-- Nouns & 77.13 & \textbf{81.80}\\
         \quad\quad\quad ~~~-- Verbs & \textbf{55.63} & 55.49\\
         \quad\quad\quad ~~~-- Adverbs & \textbf{44.44} & 42.11\\
         \quad\quad\quad ~~~-- Adjectives & \textbf{61.67} & 58.48\\
         \thickhline
         -sense & 76.91 & \textbf{78.93}\\
    \end{tabular}
\end{small}
}
\caption{$F_1$-scores of fine-grained evaluation on the PMB (v.2.2.0) 
  development set; the seq2graph models trained on \emph{gold} (left) and with \emph{gold} and \emph{silver} data combined (right) are compared.}
    \label{pmbeval}
\end{table}

\section{Conclusions}
In this paper we have introduced a novel graph parser that can
leverage the power and flexibility of sequential neural models while
still operating on graph structures. Heavy preprocessing tailored to a
specific formalism is replaced by a flexible grammar extraction method
that relies solely on the graph while yielding performance that is on
par or better than string-based approaches. Future work should focus
on extending our parser to other formalisms (AMR, MRS, etc.). We also
plan to explore modelling alternatives, such as taking different graph
generation oders into account (bottom-up vs. top-down) as well as
predicting the components of a fragment (type, number of edges, edge
labels) separately.

\section*{Acknowledgments}
We thank Yova Kementchedjhieva, Andreas Grivas and the three anonymous reviewers for their useful comments. This work was done while Federico Fancellu was a post-doctoral researcher at the University of Edinburgh. The views expressed are his own and do not necessarily represent the views of Samsung Research. We gratefully acknowledge the support of the European Research Council (Fancellu, Lapata; award number 681760, ``Translating Multiple Modalities into Text''); and the EPSRC Centre for Doctoral Training in Data Science, funded by the UK Engineering and Physical Sciences Research Council (Gilroy; grant EP/L016427/1) and the University of Edinburgh (Gilroy).

\bibliography{acl2019}
\bibliographystyle{apa}

\end{document}